\documentclass[pdflatex,sn-nature]{sn-jnl}

\usepackage{graphicx}%
\usepackage{multirow}%
\usepackage{amsmath,amssymb,amsfonts}%
\usepackage{amsthm}%
\usepackage{mathrsfs}%
\usepackage[title]{appendix}%
\usepackage{xcolor}%
\usepackage{textcomp}%
\usepackage{manyfoot}%
\usepackage{booktabs}%
\usepackage{algorithm}%
\usepackage{algorithmicx}%
\usepackage{algpseudocode}%
\usepackage{listings}%
\usepackage{tabularx}
\usepackage{tikz}
\usepackage{pdflscape}

\usepackage{todonotes}

\newif\IFproofread

\proofreadtrue

\theoremstyle{thmstyleone}%
\theoremstyle{thmstyletwo}%

\theoremstyle{thmstylethree}%

\begin{document}

\title{Argumentation-Based Explainability for Legal AI: Comparative and Regulatory Perspectives}



 \author[1]{\fnm{Andrada Iulia} \sur{Prajescu}}\email{andradaiulia.prajescu@studenti.unipd.it}

 \author*[1]{\fnm{Roberto} \sur{Confalonieri}}\email{roberto.confalonieri@unipd.it}



 \affil*[1]{\orgdiv{Department of Mathematics `Tullio Levi-Civita'}, \orgname{University of Padua}, \orgaddress{\street{via Trieste 63}, \city{Padova}, \postcode{35121}, \state{Padova}, \country{Italy}}}




\abstract{Artificial Intelligence (AI) systems are increasingly deployed in legal contexts, where their opacity raises significant challenges for fairness, accountability, and trust. The so-called ``black box problem'' undermines the legitimacy of automated decision-making, as affected individuals often lack access to meaningful explanations. In response, the field of Explainable AI (XAI) has proposed a variety of methods to enhance transparency, ranging from example-based and rule-based techniques to hybrid and argumentation-based approaches. This paper promotes computational models of arguments and their role in providing legally relevant explanations, with particular attention to their alignment with emerging regulatory frameworks such as the EU General Data Protection Regulation (GDPR) and the Artificial Intelligence Act (AIA). We analyze the strengths and limitations of different explanation strategies, evaluate their applicability to legal reasoning, and highlight how argumentation frameworks---by capturing the defeasible, contestable, and value-sensitive nature of law---offer a particularly robust foundation for explainable legal AI. Finally, we identify open challenges and research directions, including bias mitigation, empirical validation in judicial settings, and compliance with evolving ethical and legal standards, arguing that computational argumentation is best positioned to meet both technical and normative requirements of transparency in the law domain.}

\keywords{Explainable Artificial Intelligence (XAI), Computational Argumentation, Legal Reasoning, Transparency and Accountability}



\maketitle

\section{Introduction}\label{sec:introduction}

Deep learning models are known for their high performance and ability to handle complex data, but  they are considered black boxes. 
This term refers to the lack of transparency in how these models generate
outcomes based on input data: they do not naturally explain the results that they produce and it is difficult to trace how specific inputs are transformed into outputs because of the models’ complexity, which hides the internal logic to the user~\cite{IF-2023}. 

This issue is particularly problematic in the {\em legal context}, where the decisions taken by black box algorithms could have life-changing consequences for people. A recent example of the possible negative effects caused by the lack of transparency of AI models is the Dutch childcare benefits scandal, known as the {\em toeslagenaffaire}. In this case, the Dutch tax authorities used a black box algorithm to create risk profiles to spot childcare benefits fraud~\cite{heikkila2022dutch}. The model was biased and was twice more likely to flag people with double nationality as possible fraudsters.
The families suspected of fraud by the algorithm were penalized by the authorities and asked to pay back the childcare support money they had supposedly received fraudulently, at least according to the model. This incident had terrible consequences especially on families with lower incomes, however everyone that was falsely accused was put through significant emotional distress.

To avoid this, it is important and necessary to develop methods that aim to describe how black box models work and explain their decision making in a transparent way. Ensuring transparency can help maintain fairness and avoid discrimination~\cite{Almada2019}, which is often caused by inherited prejudices contained in the training data. 
Additionally, the availability of transparent technologies that provide justifications for their decisions will increase users' trust. Ultimately, users  will perceive AI decision-making as being legitimate and worthy of
acceptance~\cite{definelicht2020ai} because, given the explanation, the users will be able to accept the decision even if it is adverse~\cite{walmsley2020ai} as they understand the reason or the process behind that outcome.

Explainable AI (XAI) is a research field within AI whose focus is  to develop explainability methods and techniques that allow different AI stakeholders to understand how and why black box models make certain decisions, with the goal of promoting transparency and of mitigating biases~\cite{IF-2024,WIREs-2020}. Whilst XAI is not a new concept~\cite{WIREs-2020},  the interest in it has grown in recent years due to ethical and regulatory concerns\footnote{Under the GDPR policy, people have the right to request an explanation if they are subjected to automated decision-making systems. The regulatory framework is addressed in Section~\ref{sec:legal_frameworks}} that emerged as AI models started to be used more broadly also in sensitive domains, such as health, finance and the judicial system.  
Research in XAI has capitalized in a vast number of explainability techniques, ranging from feature-based to rule-based approaches~\cite{guidotti2018survey}, concept-based approaches~\cite{conceptXAI2023} and interpretable-by-design models~\cite{IF-2023}. 
The applicability of each approach depends on different explanation goals and target audiences~\cite{IntSys-2024}, on the accuracy-interpretability trade-off as well as the complexity of the model to be explained~\cite{IF-2023}. 
XAI is a multidisciplinary discipline~\cite{IF-2024} and it has been argued that it shares many similarities with the social sciences, particularly argumentation
theory~\cite{miller2019explanation} 

The relationship between computational argumentation, XAI and law has been explored by different authors~\cite{ijcai2021p600,Vassiliades_Bassiliades_Patkos_2021,McGregor_2023,ATKINSON2020103387,Rotolo_Sartor_2023} from different perspectives. Atkinson et al.~(2020)~\cite{ATKINSON2020103387} traced the evolution of explanation in AI and Law, arguing that approaches grounded in case-based reasoning and computational argumentation most effectively capture the contestable and value-laden nature of legal decision-making. Vassiliades et al.~(2021)~\cite{Vassiliades_Bassiliades_Patkos_2021} surveyed argumentation-based methods for XAI, showing that argumentative reasoning provides a powerful foundation for generating transparent, interactive, and persuasive explanations that reflect human deliberation and legal justification. Čyras et al.~(2021)~\cite{ijcai2021p600} formalized the landscape of argumentative XAI, illustrating how argumentation can serve not only as a reasoning model but also as an interface for contestable, user-centered explanations. Rotolo and Sartor~(2023)~\cite{Rotolo_Sartor_2023} have recently provided a systematic account of the conceptual and formal relations between argumentation, justification, and explanation in legal reasoning. 

This paper advances this line of research by positioning computational argumentation as a cornerstone for explainability in legal AI systems. It builds upon~\cite{ATKINSON2020103387} and extends these theoretical surveys by providing a comparative and applied analysis of different explanation strategies highlighting their respective advantages and limitations in legal contexts. Moreover, it situates argumentation-based explanations within the evolving European regulatory landscape, demonstrating their alignment with the transparency and contestability principles articulated in the General Data Protection Regulation (GDPR) and the forthcoming Artificial Intelligence Act (AIA). In doing so, this work argues that computational argumentation offers the most robust and normatively grounded framework for ensuring both technical transparency and legal accountability in AI-driven decision-making.


The rest of the paper is organised as follows. Section~\ref{sec:legal_frameworks} outlines the European regulatory framework relevant to explainable AI and legal reasoning. Section~\ref{sec:analysis} presents a comparative analysis of existing explanation methods, including example-based, rule-based, hybrid, and argumentation-based approaches illustrating how different forms of computational argumentation can be applied to model legal reasoning processes. Section~\ref{sec:discussion} discusses the alignment of argumentation-based explainability with current European regulatory frameworks, notably the GDPR and the AI Act. Finally, Section~\ref{conclusion} concludes the paper by summarising key insights and outlining directions for future research on integrating argumentation and explainability in AI for legal contexts.

\section{Legal Frameworks}\label{sec:legal_frameworks}

\subsection{The Right to Explanation}

Algorithms that autonomously make decisions, or even just assist users in decision-making, can greatly impact people’s lives. This creates a need for a {\em right to explanation}~\cite{kim2018informational}. Before examining the legal framework surrounding this right, it is important to note that it is first of all a moral right. The underlying principle is that as human beings, we deserve explanations when choices significantly affect us and we do not understand them~\cite{jongepier2022explanation}. The right to explanation can also be seen as a natural extension of the fundamental human rights: the use of opaque and potentially discriminatory intelligent systems violates the human right of equal treatment before the law, which is stated by the Universal Declaration of Human Rights~\cite{un1948udhr}. Providing explanations for AI-driven decisions would help prevent this and other potential violations of existing human rights~\cite{winikoff2021ai}.

In terms of the international legal framework, one of the most significant
standards is the UNESCO Recommendation on the Ethics of Artificial
Intelligence~\cite{unesco2022ethics}. As the title suggests, these are non-binding guidelines that member states may choose to follow voluntarily~\cite{alobeidi2022legal}. The Recommendation has five paragraphs---numbers 37 to 41---dedicated to the topics of explainability and transparency, emphasizing that one cannot exist without the other. Paragraph 37, for example, connects the right to a fair trial and effective remedy with the need for transparency and explainability. These rights are enabled by the possibility to challenge the decisions and, since it would be difficult to question the outcome without understanding it, this could undermine the right to a fair trial. Paragraph 38 further asserts that individuals should have access to the reasons behind decisions that affect their rights and freedoms. When AI systems are employed in judicial processes, they can certainly have a great impact on people’s future or daily life, so there are the means for a right to explanation to be granted.

Another key international source of guidelines for trustworthy AI is OECD
Recommendation of the Council on Artificial Intelligence~\cite{oecd2024ai}. This document, updated in May 2024, includes a dedicated section (no. 1.3) on
transparency and explainability. It emphasizes that AI actors should provide meaningful information in order to enable those affected by the AI system to understand the outputs and, if necessary, to challenge them. However, like the UNESCO recommendations, these guidelines are not legally binding, which limits their practical effectiveness.

There are stronger legal instruments within the European Union: the General Data Protection Regulation (GDPR), which is legally binding across member states, provides an implicit right to explanation, and the recent adoption by the European Parliament of the Artificial Intelligence Act (AIA) marks a significant development because this new regulatory framework is the first in the world which applies specifically to Artificial Intelligence. Combined, these European regulations have a profound impact not only within the EU but also on global AI governance standards, ensuring individuals' rights in the context of automated decision-making. It will become clear how these regulations grant the right to explanation as we explore their specifics in the next sections.

\subsection{General Data Protection Regulation (GDPR)}

The General Data Protection Regulation (GDPR)~\cite{eu2016gdpr} is designed to protect the privacy and other rights of individuals that could be put at risk by improper data processing techniques. Although the existence of a right to explanation is not explicitly stated in the GDPR, several of its articles do require for meaningful information to be provided to individuals who are subjected to decisions made by algorithms~\cite{pak2022responsible}. If automated decision-making has a legal effect or another significant impact on a subject, Article 22 limits its application by allowing it only if the subject explicitly consents or if other specific conditions apply, such as a contract or the approval of a member state. Even when automated decisions are authorized, the subject has the right to request human intervention and challenge the decision, suggesting an implicit right to explanation.

It is explicitly stated in articles 13, 14, and 15 that ``meaningful information about the logic involved'' should be provided to the subject in order to ensure transparency in automated decision-making. The requirement that the information must be meaningful guarantees that the explanations provided are not just available to the subject, but also understandable, providing individuals with the grounds to challenge the outcome if necessary. Despite this, the GDPR was not created with AI in mind, considering that its primary goal is data protection and seeing that the right to explanation is implicit. Because of this, the application of its principles is subject to some limitations and ambiguities, so the new AI Act is crucial to establish a safer and more transparent legal environment by providing specific regulations that also address fundamental concerns like the algorithmic bias and the accountability of AI systems.

\subsection{Artificial Intelligence Act (AIA)}

The Artificial Intelligence Act (AIA)~\cite{eu2024aiact} was published in the Official Journal of the European Union in June 2024, and subsequently the European Artificial Intelligence Office was established to ensure the Act’s effective implementation across EU member states and to provide legal certainty. This is a big step forward in contrast with the previous regulatory approaches that focused more on ethics and relied on the self-regulation of AI systems, leading to ambiguity and undermining public trust. The provisions of the AI Act, which will gradually come into force over the next years, aim to balance the rapid expansion of the AI industry with the protection of citizens' fundamental rights by encouraging the development of safe and reliable AI systems. The AIA also introduces a distinction of AI models into four categories of alleged risk based on their intended use: unacceptable risk, high risk, limited risk, and minimal risk. 

In chapter 3 of the Act, articles 9 to 15 outline various requirements that are deemed as essential for ``high-risk'' systems to be safely deployed in the market, which will be mandatory for AI systems used in the legal context as they are part of the high-risk category. These provisions include the implementation of a risk management system that must be updated throughout the AI system’s lifecycle; as well as requirements about the data used in training, validation and testing, which must be free from biases, relevant and representative of the context in which the AI system will be used. Other important requirements address transparency and explainability: high-risk AI models must be accompanied by instructions that explain their capabilities and limitations and enable users to interpret their outputs. Additionally, these systems must be designed to ensure that human operators can understand their workings, promoting transparency by design rather than relying on post-hoc explainability. High-risk AI systems are also subject to the CE marking: only the conforming models that meet the essential requirements can be freely imported and distributed throughout the EU~\cite{edwards2022explainer}. 

Another significant provision of the AIA is article 86, which explicitly guarantees individuals the right to be given clear and meaningful explanations of decisions made by high-risk artificial intelligence systems that have significant consequences on their life, including legal effects. However, in order to give high-risk systems more time to comply, this provision will be one of the last to come into force on August 2, 2027, so until then the GDPR remains an important tool for granting explainability.

As AI continues to evolve rapidly and reach a broader audience, existing legal frameworks struggle to keep pace, often rendering them outdated and insufficient to ensure public safety and a proper understanding of these emerging technologies. The AIA represents a significant advancement that will set the standards for AI governance not only within the European Union, but also globally~\cite{cancelaouteda2024euai}.

\section{Computational Argumentation and Explainability Solutions}\label{sec:analysis}

Building upon the legal and ethical foundations outlined in the previous section---where transparency, accountability, and human oversight were identified as essential requirements---this section explores how computational models can operationalize these normative principles. In particular, it focuses on computational argumentation as a paradigm capable of providing structured, interpretable, and normatively grounded explanations. By modeling reasoning as a process of constructing and evaluating arguments, computational argumentation offers a means to reconcile the technical goals of XAI with the legal demand for contestable and justifiable decision-making.

\subsection{Example-based Explanations}

The main idea behind example-based explanations, also referred to as case-based, is that similar cases should be treated in a similar manner. In other words, past decisions establish precedents that must be followed unless the current case stands out in a significant way. 
In common law, lawyers also typically argue by citing precedent cases, using both positive examples (cases with similar outcomes) and negative ones (cases with similar aspects but different outcomes) to build contrastive explanations~\cite{miller2019explanation}. 

Counterfactual explanations highlight what minimal change in the input would have led to a different outcome. If the contrast case is not explicitly provided, the system typically selects the most similar precedent from the case database, which corresponds to the output that is caused by the minimal change of the original input. In other words, counterfactual explanations answer the question ``what would need to be different to achieve the expected outcome?'' by highlighting the key distinguishing factors between the two cases.

Contrastive explanations are very intuitive to humans because they clarify why an event occurred instead of an alternative one~\cite{jacovi2021contrastive}. Ideally, an explanation should address why one outcome was reached over another, instead of simply presenting examples of similar cases. The challenge that XAI has to face is to identify the other hypothetical event, as it is often implicit. In fact, when a person asks ``why did X happen?'' they  typically imply ``why did X happen instead of Y?'', where Y represents an event that did not occur but was anticipated by the user, since people rarely ask for explanations when outcomes align with their expectations.

\begin{figure}[!t]
    \centering
    \includegraphics[width=0.7\linewidth]{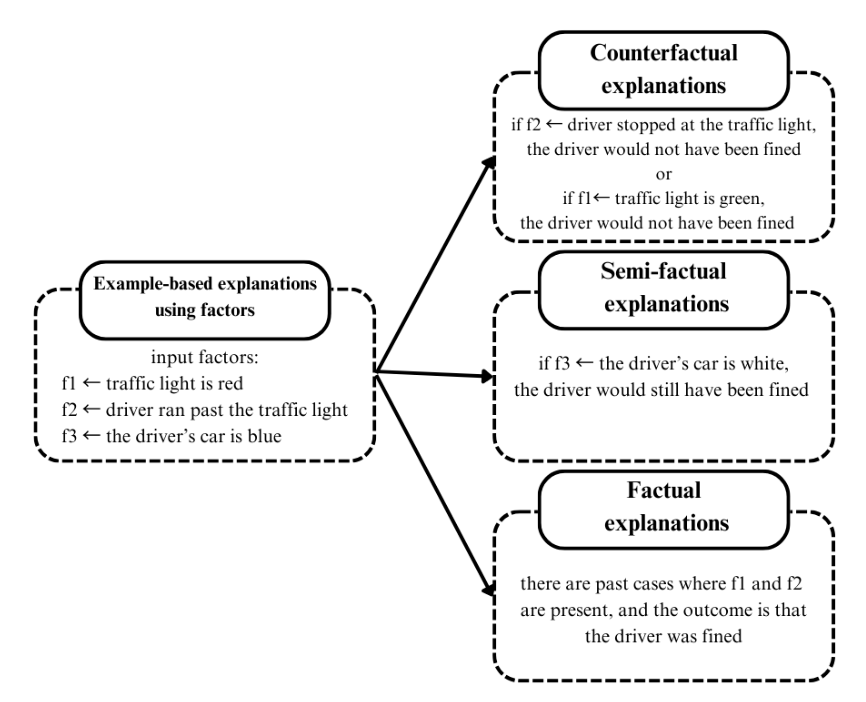}
    \caption{Example of counterfactual, semi-factual and factual explanations.}
    \label{fig:example-based-explanations}
\end{figure}

In order to replicate this type of explanation with an AI model in the legal system, precedents are represented as sets of facts that determine what makes one case different from another, specifically, which facts led to a specific outcome. To determine this causality, it is necessary to understand which factors are truly responsible for the outcome and which ones are irrelevant~\cite{robeer2018contrastive}. For example, if a fact appears in both the actual output and the expected output, it is not considered a causal factor. The model does not need to consider all the possible causes of an event~\cite{miller2021contrastive}, but only focus on those relevant to the comparison case since humans also naturally provide partial explanations, focusing on key factors that influenced an outcome rather than listing all possible causes. Both relevant and irrelevant facts can be used in explanations: relevant facts are used for counterfactual explanations, while irrelevant ones contribute to semi-factual or factual explanations. 

These three types of explanations are illustrated in Fig.~\ref{fig:example-based-explanations} with a simple example of a driver not stopping at a red light and being fined as a result. Counterfactual explanations highlight what minimal change in the input would have led to a different outcome. If the contrast case is not explicitly provided, the system typically selects the most similar precedent from the case database, which corresponds to the output that is caused by the minimal change of the original input~\cite{stepin2021survey}. In other words, counterfactual explanations answer the question ``what would need to be different to achieve the expected outcome?'' by highlighting the key distinguishing factors between the two cases. On the other hand, semi-factual explanations identify facts that do not influence the outcome, so even if these facts had been different the result would have remained the same. Meanwhile factual explanations are the simplest form of example-based reasoning, simply providing past cases with similar inputs and identical outcomes~\cite{artelt2023idk}.

\begin{figure}[!t]
    \centering
    \includegraphics[width=0.7\linewidth]{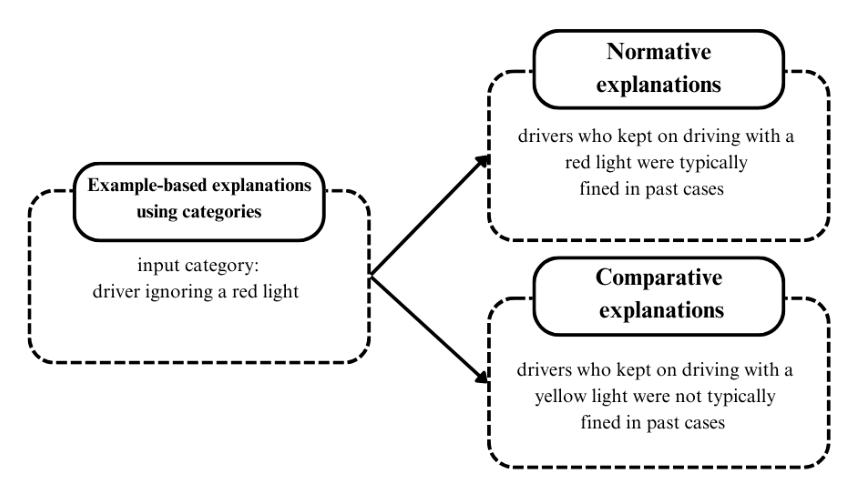}
    \caption{Example of normative and comparative explanations.}
    \label{fig:normative-explanations}
\end{figure}

Another approach to example-based explanations is shown in Fig.~\ref{fig:normative-explanations}. It involves categorizing cases from the training set into different classes based on shared characteristics. With this framework the model can generate two types of explanations: normative and comparative. Normative explanations provide examples from the same category as the given case, so they are similar to factual explanations but without analyzing individual factors separately. Comparative explanations present examples from different but similar classes, which resemble counterfactual explanations but, since the factors are not examined one by one, the most appropriate contrast case may not always be selected, potentially leading to unexpected examples that confuse users~\cite{park2022examples}.

\subsection{Rule-based Explanations}

Explanations using rules are particularly well-suited for civil law traditions but can also be applied in common law systems. Since the latter operates through analogical reasoning, we can formalize explicit rules by extracting and applying legal principles from past cases, resulting in creating an abstract representation of the underlying legal principles. However, representing cases in a formalized way in order to allow a system to extract a rule is a complex task~\cite{blass2025analogy}.

Rule-based contrastive explanations, such as ``if ... then...'' statements, explicitly define the decision boundary between the given output and the alternative outcome. This can improve the user’s understanding of the system behavior by clearly expressing the distinguishing factors behind a decision, which help individuals to correctly identify the factors that played a decisive role in the output~\cite{vanderwaa2021evaluating}. In contrast, example-based explanations require users to infer the underlying rule from counterfactual examples.

One notable method in this category is LORE (Local Rule-Based Explanations)~\cite{guidotti2018local}, an agnostic technique that provides insight into a system’s decision-making process for specific instances by examining the input-output behavior of the system without requiring knowledge of its internal workings. It generates explanations consisting of a decision rule, which clarifies the reasoning behind a decision, and a set of counterfactual rules, which suggest modifications to the input features that would lead to a different outcome~\cite{guidotti2018local}.

\begin{figure}[!t]
    \centering
    \includegraphics[width=0.7\linewidth]{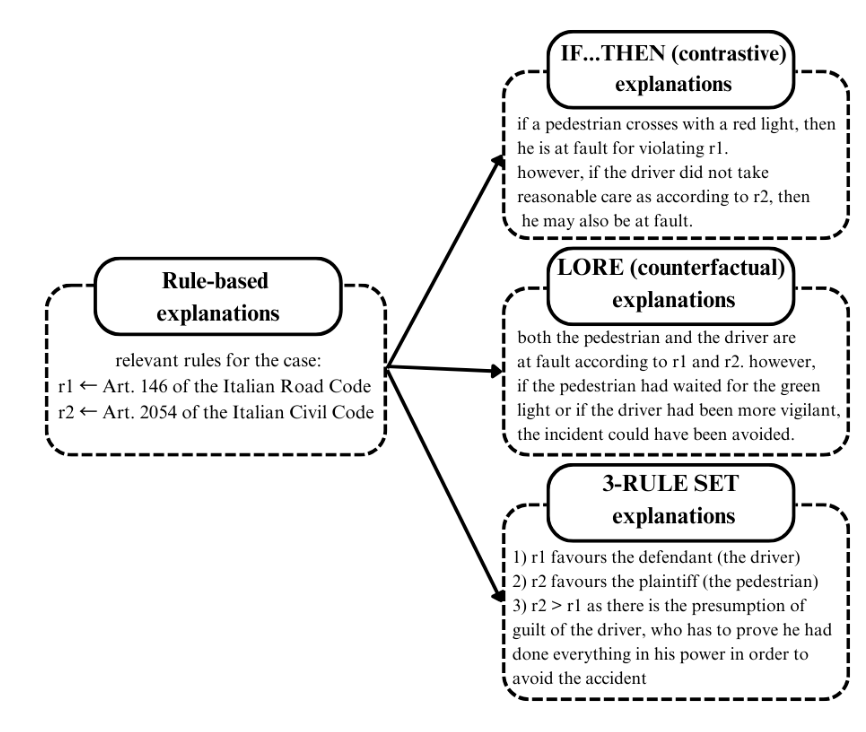}
    \caption{Example of the main methods of rule-based explanations.}
    \label{fig:rule-based-explanations}
\end{figure}

In legal reasoning, precedents can also be structured as a set of three rules: two defeasible rules, one that favors the plaintiff and one that favors the defendant, and a priority rule which establishes which rule takes precedence in case of conflict. This and the other two main rule-based methods described above are illustrated in Fig.~\ref{fig:rule-based-explanations} with a slightly different example than the one in the previous section, where a pedestrian (the plaintiff) crosses the street with a red light and is hit by a driver (the defendant) who had a green light. 

However, the three-rules technique has some limitations: if no priority rule exists, it cannot resolve conflicts between rules that have not previously been decided on. Moreover, legislative gaps present another challenge: new cases may not be covered by any existing rule, making it difficult to even determine which factors are relevant if they have never been considered before. While legislative gaps are hard to solve, one possible solution for the absence of the priority rule is to establish preferences between legal values. However, this approach must also take into account that values shift over time, and legal systems evolve through new legislation or precedents that diverge from past sentences~\cite{atkinson2019reasoning}.

Methods for extracting rules seek to balance interpretability and accuracy as
there is a trade-off between these two attributes: the ideal rule-set should be as concise as possible, minimizing both the number of rules and their length in order to be easily understood by end-users, but rules that are too simple or short may reduce the accuracy and reliability of the explanations~\cite{vilone2020comparative}. Finding the right balance between clarity and precision is a key challenge in rule-based legal reasoning.

\subsection{Hybrid Explanations}

Hybrid explanation techniques aim to integrate the strengths of example-based
and rule-based approaches: as we have seen in the previous two sections, example-based explanations provide intuitive justifications through reference to precedents, enabling users to comprehend decisions through analogical reasoning, but with the drawback of requiring individuals to infer the principles of the decision-making process themselves. We have also seen that, in contrast, rule-based explanations openly state how and which legal factors influenced the outcome, even though they also have a big disadvantage as they fail to solve cases that do not fit into existing and defined principles. 

A hybrid approach can bridge these gaps by combining explicit legal rules with illustrative precedent cases, providing both a structured logical framework and real-world examples. This approach improves transparency by allowing users to both see the reasoning process behind a decision and compare it to actual past cases~\cite{atkinson2020explanation}. Despite their potential, hybrid explanation systems are not very widely researched in explainable AI for the law domain, perhaps because of the difficulty of creating a seamless integration of two different techniques.

\subsection{Argumentation-based Explanations}

Arguments serve as justifications for adopting a particular conclusion and they are fundamental in the legal field, where both parties in a trial present arguments to support their claims. Toulmin’s model is one of the most widely used techniques for assembling arguments, whose structure is intuitive for non-logicians such as lawyers and jurors~\cite{toulmin1958argument,toulmin2003argument}. This model divides arguments into six distinct components that play different roles:

\begin{enumerate}
    \item the {\em claim}, which is the argument’s conclusion and states a fact or an event
    \item the {\em qualifier}, which expresses the degree of confidence of the claim;
    \item the {\em premises}, which constitute the facts supporting the claim;
    \item the {\em warrant}, which is the inference rule that links the premises to the claim;
    \item the {\em backing}, which explains the warrant by citing a general rule, such as a statute law;
    \item the {\em rebuttal}, which constitutes exceptions, that are reasons about why the claim might be false or the warrant inapplicable.
\end{enumerate}

\begin{figure}[t!]
    \centering
    \includegraphics[width=0.75\linewidth]{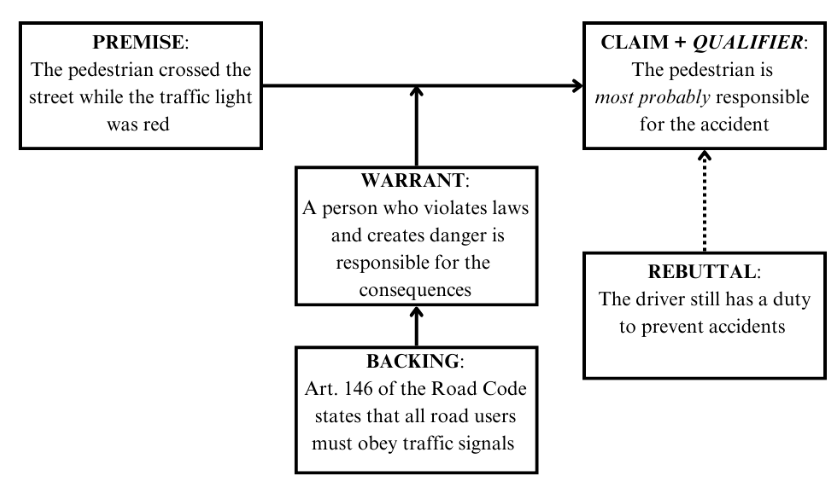}
    \caption{Example of the structure of an argument according to Toulmin’s model~\cite{toulmin1958argument,toulmin2003argument}.}
    \label{fig:toulmin}
\end{figure}

In legal explanations, arguments typically follow a similarly structured order: first presenting the fact, followed by the backing (which reinforces the fact), and concluding with the rebuttal (which challenges the fact). Qualifiers are often implicit as normally people do not state the probability of their claim, but they sometimes use expressions like ``certainly'' or ``most probably''. The warrant is also generally implicit since it is part of the general knowledge shared by the participants in the debate. Fig.~\ref{fig:toulmin} shows the structure of an argument with the same example as the previous section, where a pedestrian crossed the street with a red light which resulted in an accident with a driver who had a green light.

Another important feature of argumentation is that it is defeasible, which
means that arguments can be challenged and reassessed over time. This is because argumentation is not merely about finding out if a belief is true but instead it is about what should be done in a given situation. Thus, it is always possible to express some form of doubt about a claim. This represents practical reasoning, which allows multiple valid perspectives and does not imply that one of the two parties is right and the other one is wrong, as everyone has different interests and priorities and might be right according to their own perspective. The purpose of an argument, therefore, is to persuade rather than to establish an absolute truth. The persuasiveness of an argument is subjective, since the same case might convince some people but fail to convince others~\cite{benchcapon2007argumentation}.

In the legal domain, disputes often revolve around two types of burden: the
burden of production, which is that of providing evidence, and the burden of
persuasion, which is about convincing the judges. In civil law, the plaintiff has both burdens relatively to the facts that support their claim and the defendant is responsible for providing the exceptions. For example, a plaintiff who asks for compensation has to prove that the defendant caused harm and that the harm was caused by intentional or negligent actions. However, in certain cases the burden shifts and the plaintiff only needs to prove the harm, while the defendant has to provide a convincing argument that they were diligent and not negligent~\cite{calegari2021burden}. 

Argumentation Frameworks (AFs) provide a formal structure for organizing and evaluating arguments and the relations between them. Dung introduced the first type of AF which is now known as {\em Abstract Argumentation Framework} (AAF)~\cite{dung1995acceptability}. To decide whether an argument is accepted or rejected AAFs do not consider the arguments’ internal structures, but only the attack relationships which encode the existing conflicts between them. Hence, the ``abstract'' in the name, but it is still useful to represent arguments using Toulmin’s model so that their individual validity is verifiable by checking the premise, the warrant and the backing.

In general, attacks on a claim consist of counterarguments that justify the negation of the claim and, within this framework, an argument is considered acceptable only if none of its attackers are accepted. Dung proposed four types of formal methods for evaluating sets of arguments, called argumentation semantics~\cite{baroni2009semantics}: {\em complete}, {\em grounded}, {\em preferred}, and {\em stable}. They create different kinds of subsets, named {\em extensions}, of claims that are collectively acceptable and follow the conflict-free principle, so arguments that attack each other cannot be included in the same extension. 

Another global condition for these subsets is that of reinstatement, which means that the arguments in the subset must attack back all arguments that attack them, so that the extension can survive. Complete extensions include an initial set of arguments and all the arguments they progressively defend, while grounded extensions include an initial set of arguments and the arguments that are defended by the initial arguments. Stable extensions are a special type of complete extensions which are able to attack all arguments that are not included in them, and preferred extensions are also a subtype of complete extensions that are as large as possible~\cite{baroni2009semantics}. 

The use of argumentation schemes can better capture the reasoning in reaching a decision by explaining not only in terms of previous cases, but in terms of rationales for those decisions. There are different types of argumentation schemes, some of the most common being argument from expert opinion, from popular opinion, from analogy, from cause to effect, etc. Each one of them is associated with critical questions that raise doubt about the acceptability of the argument to evaluate its strength~\cite{walton2009argumentation}. It is important, but also problematic, to identify the correct scheme that is being used so the appropriate critical questions can be asked.

Argumentation frameworks and argumentation schemes have complementary purposes: the first is a general framework for analyzing individual arguments, the seconds are a formal model that represents arguments and their relations, and the thirds are general patterns of reasoning for different types of arguments. AFs can be visualized as a directed graph, where the nodes are the arguments proposed and the edges represent attack relationships. If one argument attacks another, it can defeat or weaken it depending on their relative strengths. AFs can identify cycles in graphs, allowing courts to find out the source of the disagreement and assess the consequences of adopting different legal positions~\cite{habernal2017structure}. 

\begin{figure}[!t]
    \centering
    \includegraphics[width=0.6\linewidth]{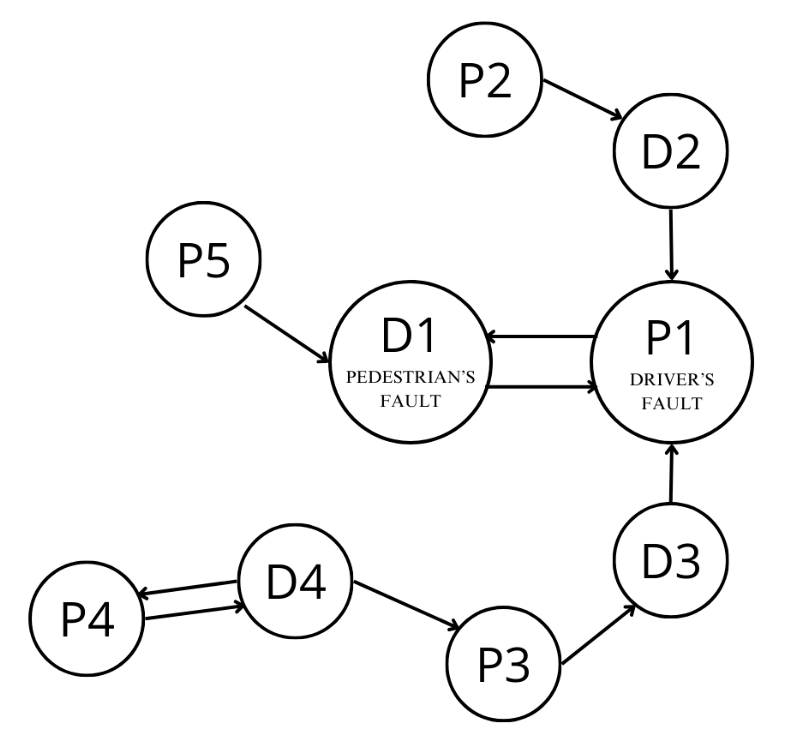}
    \caption{Example of an argumentation graph.}
    \label{fig:aaf}
\end{figure}

Fig.~\ref{fig:aaf} shows an argumentation graph of the previous example, with the pedestrian’s claim being ``P1: the driver is at fault'', and the driver’s competing claim being ``D1: the accident is the pedestrian’s fault''. They attack each other, but there is not enough knowledge to accept one of the claims and reject the other, so the two parties have to present other arguments to weaken the opposing claim, for example:
\begin{itemize}
    \item $D2$: the pedestrian crossed with a red light. (Attacks $P1$)
    \item $P2$: the pedestrian is a child who does not understand traffic rules. (Attacks $D2$)
    \item $D3$: the pedestrian unpredictably ran into the street. (Attacks $P1$)
    \item $P3$: the driver had enough time to stop. (Attacks $D3$)
    \item $D4$: the driver reacted as best as he could. (Attacks $P3$)
    \item $P4$: the driver was distracted. (Attacks $D4$) ($D4$ attacks $P4$)
    \item $P5$: the driver was speeding. (Attacks $D1$)
    \item The attack relations between the arguments is defined as: $R = \{(D2,P1), (P2,D2), (D3,P1), (P3,D3),(D4,P3), (D4,P4), (P4,D4), (P5,D1)\}$
\end{itemize}

The graph allows to immediately identify the main sources of the disagreement as there are two cycles $(D1-P1)$ and $(D4-P4)$. Obviously, the main claims ($D1-P1$) started the disagreement, but it is interesting to notice that the two parties also disagree about the diligence or negligence of the driver ($D4-P4$). In this case, it is the driver that has to prove somehow that he did everything he could to avoid the accident, and if he cannot prove it then the $D4$ argument is defeated. Using Toulmin’s scheme for representing arguments could prevent these cycles as the argument would have not been accepted from the start if it lacked a backing
strong enough to justify it. 

Preferred semantics are often used for legal reasoning because it selects the strongest position among the two parties, as it coincides with the largest complete extension. The preferred semantics in this example leads to the set $\{P1, P2, P3, P4, P5\}$, so the pedestrian wins the case and the final decision is that the accident was mostly the driver’s fault. Another conflict-free and complete set is $\{D1, D2, D3, D4\}$, but this set is smaller than the previous so it is not selected by the preferred semantics. If the $D4$ argument was rejected because the driver could not
prove his diligence, then this set would have been even smaller. For a hypothetic application of an AI system in a court case, some arguments could be presented at the beginning by the two parties and become part of the knowledge base, then the parties can present counterarguments based on the initial arguments presented by the opposing party, which could be added to the graph at a later time and the outcome could change dynamically, if the new arguments are strong enough to overturn the decision. 

AAFs only include attack relationships between two arguments, but the interaction between arguments is more complex: an argument can also be attacked by critical questions, or it can be supported by other arguments. In order to overcome the limitations of the original Abstract Argumentation Framework, some extensions have been proposed. A slightly more complex system is that of Bipolar Argumentation Frameworks (BAFs), which consider support relations in addition to attack ones. This framework is more expressive as it can represent a broader pool of interactions between arguments~\cite{cayrol2009bipolar}. 

Value-based Argumentation Frameworks (VAFs) are another extension of AAFs which is very interesting for the legal context as it considers the values of the audience, which can be viewed as the court, when evaluating arguments. When an argument attacks another, and both arguments are valid, the outcome of the attack depends on the values that the arguments pursue and how important these values are considered by the court~\cite{benchcapon2009abstract}. The preferred argument is the one that promotes the value ranked higher by the court. 

Abstract Dialectical Frameworks (ADFs)~\cite{brewka2010adf} generalize AAFs by introducing an acceptance condition associated to each argument that allows representations of different types of relations, which include but are not limited to support and attack, as well as different types of arguments~\cite{brewka2017adf}. For example, it can represent a joint attack of two arguments as an acceptance condition for a counterargument to be rejected, so that a singular attack is not enough to defeat the counterargument, which is perceived as stronger. In a legal context, acceptance conditions can be seen as a knowledge base~\cite{atkinson2020explanation} that specifies under what conditions a certain argument should be accepted or rejected. For instance, in the previous example with the driver and the pedestrian, the argument about the diligence of the driver ($D4$) could not be able to defeat alone the counterargument about the driver having enough time to stop ($P3$), but would also need for the visibility conditions of the road to be scarce (new argument $D5$). This constraint could be mapped as an acceptance condition for $P3$ to be rejected, so that if only one argument between $D4$ and $D5$ is present, $P3$ still survives.

To better understand how Argumentation Frameworks can be applied to
modelling legal cases, and to comprehend the advantages of the new proposed
extensions compared to Dung’s first Argumentation Framework, each of the next sections presents a simple legal example for each AF reviewed in this section.

\subsubsection{Abstract Argumentation}

The first example to be examined will feature once again the original Abstract Argumentation Framework proposed by Dung~\cite{dung1995acceptability}, but in a different legal context from the example in the previous section to prove that AAFs are very flexible, despite their simplicity. This case example revolves around a copyright infringement issue and intellectual property. A blogger Bob posted an
article with some extracts from a newly released book, and the publishing house Hero alleges that Bob has violated copyright law. In the Italian legal context, art. 70 of the Copyright Law regulates the fair use of someone else’s creation, which is permitted if the author and the title of the original work are cited, and if it used for discussion or criticism purposes. Art. 158 of the Copyright Law is also relevant in this example as it administers economic loss of the author consequent to improper copyright usage.

This example could be mapped into the abstract argumentation framework as follows:
\begin{itemize}
 \item $H1$: Bob unlawfully used excerpts from the book in his blog.
 \item $B1$: Bob’s use of the excerpts from the book adheres to the law.
 \item $B2$: Bob mentioned the title of the book and the author in the article.
 \item $B3$: Bob used brief extracts from the book for criticism purposes.
 \item $H2$: The excerpts used by Bob contained the main plot points of the book.
 \item $H3$: The article discouraged book purchases, creating economic competition.
 \item $B4$: Bob did not pursue commercial goals with his article.
 \item $H4$: The article caused a drop in sales of the book, resulting in economic loss.
 \item $H5$: Bob has ads on his blog, so he profited off the article.
\end{itemize}

The attack relations between arguments are defined as:
\begin{itemize}  
    \item $R = \{(B1,H1), (H1,B1), (B2,H1), (B3,H1), (H2,B3), (H3,B1), (B4,H3),$ $ (H4,B4), (H5,B4)\}$
\end{itemize}

\begin{figure}[!t]
    \centering
    \includegraphics[width=0.7\linewidth]{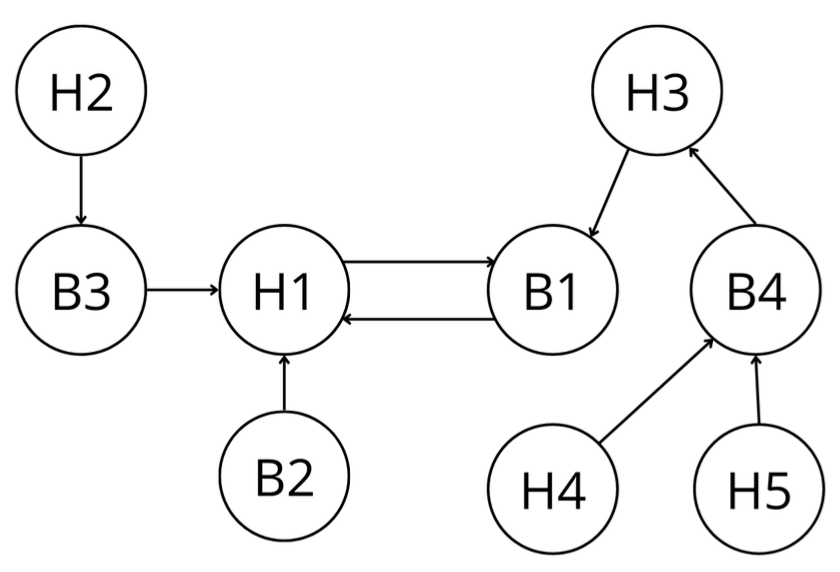}
    \caption{Graph of the Abstract Argumentation Framework used in the first example.}
    \label{fig:aaf2}
\end{figure}

Fig.~\ref{fig:aaf2} represents the Abstract Argumentation Framework graph of the example, we can see that the conflict originates from the claims $H1$ and $B1$. By applying the preferred semantics, we obtain the subset $S = \{H1, H2, H3, H4, H5 \}$ which defeats all other arguments. Thus, in this scenario the publishing house Hero wins as Bob unlawfully used copyrighted material to profit and caused economic damage to the author. Another admissible extension is $A = \{B1, B2, B3, B4 \}$ which supports the claim that Bob acted by the law, but it is not the  preferred extension as it is not the maximal admissible set. 

\subsubsection{Bipolar Argumentation}

For the second example, we will use Bipolar Argumentation Frameworks (BAFs)~\cite{cayrol2009bipolar} to model the arguments of the dispute, which add support relations on top of attacks. This case example is about a contract breach, which is a recurring problematic situation in civil law, where company Cube hires freelancer Francis to develop a software. However, when the final product is delivered, Cube refuses to pay the full amount because they claim that the software doesn’t meet the agreed criteria. Some relevant articles about this issue are present in the Italian Civil Code, including: art. 1218 which determines when a party is liable for not performing the agreed task, art. 1374 which states that the execution of the contract must be performed in good faith, art. 1455 which establishes that minor breaches do not justify contract termination, and art. 2226 which regulates defects of the performance agreed in the contract.

The bipolar argumentation of the case could be mapped as:
\begin{itemize}
    \item $C1$: Francis did not fully deliver the contract so the payment should be reduced.
    \item $F1$: The contract was fulfilled within reasonable expectations and Francis deserves full payment.
    \item $C2$: The software does not meet all agreed criteria.
    \item $F2$: The software meets industry standards.
    \item $F3$: The contract terms were ambiguous about the performance expectations.
    \item $C3$: Francis exploited the loose terms of the contract to deliver a bad software.
    \item $F4$: The contract did not specify penalties for non-conformities.
    \item $C4$: The software has bugs that prevent proper functioning.
    \item $C5$: Cube invested extra resources for fixing the software.
\end{itemize}

The attack and support relations are, respectively:
\begin{itemize}
    \item $R_a = \{(F1,C1), (C1,F1), (F3,C2), (C3,F1), (F4,C1), (C4,F1)\}$
    \item $R_s = \{(C2,C1), (F2,F1), (C3,F3), (F4,C2), (C4,C1), (C5,C4)\}$
\end{itemize}

\begin{figure}[!t]
    \centering
    \includegraphics[width=0.7\linewidth]{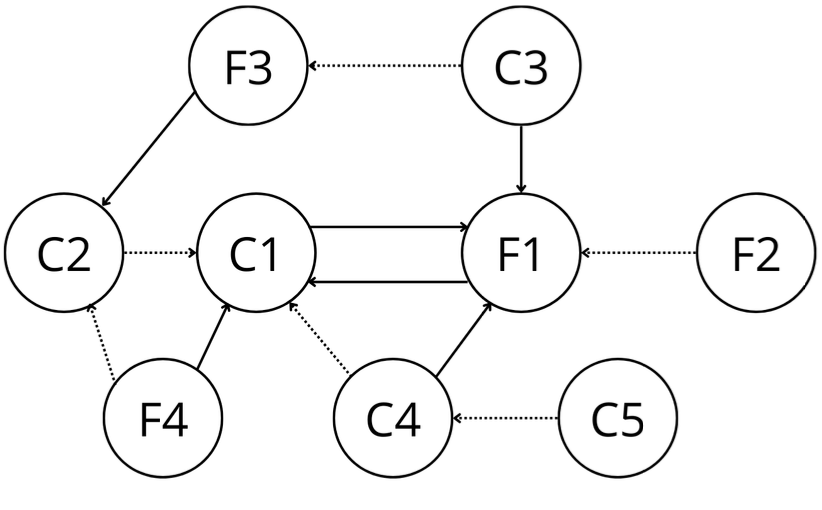}
    \caption{Graph of the Bipolar Argumentation Framework used in the second
example.}
    \label{fig:baf}
\end{figure}

Fig.~\ref{fig:baf} shows the BAF directed graph of this example, where the attack relations are represented by solid line arcs and the support relations by dotted line arcs. The only cycle in the graph is $(C1-F1)$ which identifies the main conflict between the two parties about whether the contract was fulfilled, and full payment is expected. 

In bipolar complete semantics, an argument is justified either if it is not defeated by attackers, or if it is defended by at least one justified supporter. In the example, $F1$ (Francis deserves full payment) is justified if its supporter, $F2$, is justified, and since it has no attackers we can suppose it is. But $C1$ (the payment should be reduced) is also justified because it is supported by $C4$, which is supported by $C5$, so this argument could be perceived as stronger than $F1$. $C1$ is also supported by $C2$ (the software does not meet the agreed criteria), which is
supported by $F4$ because Francis implicitly confirms that non-conformities might exist by claiming that the contract did not specify penalties for this scenario, and is attacked by $F3$, that is supported by $C3$ because Cube implicitly validates that the contract was vague by claiming that Francis exploited the ambiguous terms. Since $C3$ is not attacked by other arguments, we can suppose it is justified, so $F3$ is defended and $C2$ is defeated, making $C1$ lose a supporter.

We can conclude that $C1$ has one strong supporter ($C4$ supported by $C5$) and one attacker ($F4$), while $F1$ has one supporter ($F2$), one attacker ($C3$) and one strong attacker ($C4$ supported by $C5$). The outcome of this argumentation is that the company Cube has the right to reduce the payment because the software that freelancer Francis delivered had major bugs to the point that the company needed to get it fixed by someone else, paying extra.

\subsubsection{Value-based Argumentation}

The third example will be modeled using Value-based Argumentation
Frameworks (VAFs)~\cite{benchcapon2009abstract}, which map a value to each argument and in this case only support attack relations for simplicity. 

The example features a customer Chris who demands a refund to his bank because of an unauthorized payment. The bank first refunds him, as required by the law in the EU, but then demands payment back arguing that the transaction was approved using proper authentication, since no suspicious activity was found upon internal investigation. Chris then sues the bank, claiming that his account was hacked. As previously mentioned, in the EU banks are required to refund customers for unauthorized transactions within 24 hours according to the Payment Services Directive (PSD2), which also demands banks to adopt strong customer authentication (SCA) methods for online payments. These conditions are also found in art. 62 of the Italian Consumer Code. Another relevant regulation in this context is art. 7 of the Legislative Order 11/2010, which provides some obligations for the customer by requiring that he adopts all reasonable measures to protect his security credentials and payment instruments.

The VAF of the case example could be mapped as:
\begin{itemize}
    \item $C1$: Chris deserves a full refund from the bank. value: {\em customer protection}
    \item $B1$: The bank is not liable for a full refund. value: {\em fairness}
    \item $C2$: The bank failed to protect the customer’s account. value: {\em customer protection}
    \item $B2$: The bank notified the customer of potential phishing risks multiple times. value: {\em compliance to law}
    \item $C3$: The bank’s security measures were not strong enough. value: {\em security of the system}
    \item B3: The transaction was authenticated correctly. value: {\em fairness}
    \item $C4$: The bank failed to recognize that the transaction was suspicious. value: {\em security of the system}
    \item $B4$: The customer shared credentials negligently. value: {\em due diligence}
    \item $C5$: The customer’s account was hacked. value: {\em customer protection}
\end{itemize}

The attack and relations are:
\begin{itemize}
    \item $R = \{(C1,B1), (B1,C1), (C2,B1), (B2,C2), (C3,B1), (B3,C1), (C4,B3),$ $(B4,C1), (C5,B4), (B4,C5)\}$
\end{itemize}

\begin{figure}[!t]
    \centering
    \includegraphics[width=0.7\linewidth]{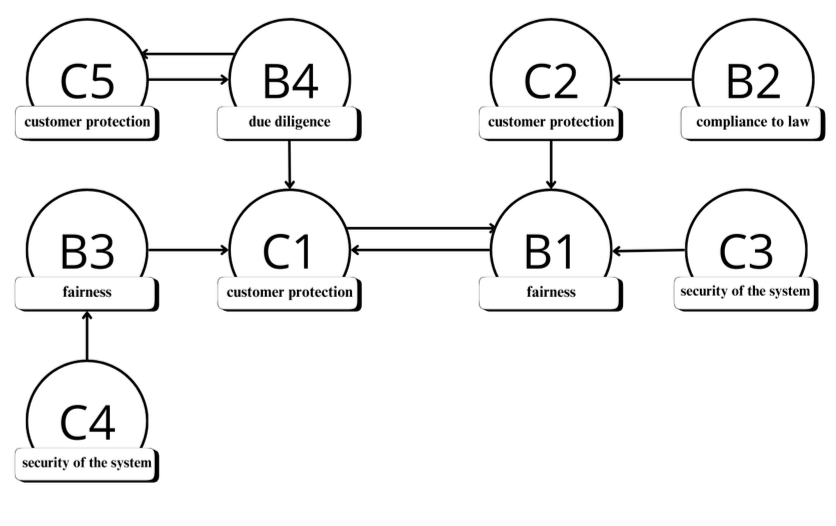}
    \caption{Graph of the Value-based Argumentation Framework used in the third example.}
    \label{fig:vaf}
\end{figure}

Fig.~\ref{fig:vaf} represents the graph of the VAF for this example, where every argument is associated with the value it promotes, and the attack relations are identified by directed arcs. The graph clearly shows four main conflicting arguments that form two cycles: the first is, as usual, constituted by the initial claims $C1-B1$ which originated the dispute about whether the bank is liable for a refund, and the second displays the disagreement between $B4-C5$ concerning the negligence of the customer.

A preference relation (e.g., a strict total order $\succ$) among the values is needed to correctly apply semantics for VAFs, based on what courts typically prioritize in disputes. It is important to notice that this preference relation is not fixed as different courts might prioritize different values, and values might even change over time. For our example, let us suppose the court order the values as following:
\begin{enumerate}
    \item customer protection  
    \item compliance to law 
    \item fairness
    \item security of the system
    \item due diligence
\end{enumerate}

While examining the attack relations, we can assess how heavy the weights of the arguments are by looking at the scale of values and determine how strong is its attack. The most important value is customer protection, since even if the bank complied to the law and met all requirements, the customer must still be refunded immediately in case of unauthorized payments. The customer’s main argument $C1$ (about deserving a full refund) pursues this value, so it is already considered stronger than its main opponent $B1$, but we should still evaluate how the other arguments play into role. 

Under preferred semantics the set of arguments $\{C1, C2, C5\}$ win against
all their attackers as they pursue the highest ranked value. The other claims that defend Chris, $C3$ and $C4$ which pursue the security of the bank system, are defeated by $B1$ and $B3$ respectively, which pursue fairness, but the admissible extension favoring the bank is constituted only by $\{B3\}$ since $B1$ is defeated by $C1$ and $C2$, so the customer wins in this case. Probably, the only way for the bank to win would be to prove that the customer’s negligence was gross or that the customer committed fraud, so that he is not worthy of protection anymore. 

It is interesting to see that this framework allows to see the preferred party since the very beginning, without even considering all the arguments yet. If the other party provided appropriate and powerful counterarguments they could still change the final outcome, but they would know from the start that the court is inclined to be on the side of the opponent and that they would need very strong evidence to persuade them.

\subsubsection{Abstract Dialectical Argumentation}

The fourth and final example is a case of criminal law and will be modeled
using Abstract Dialectical Frameworks (ADFs)~\cite{brewka2010adf}, which introduce an acceptance condition associated to each argument allowing representations of different types of relations, including support and attack like BAFs but through logical dependencies. 

This example is about a controversy regarding self-defense, where Adam is accused of assaulting Bob, but Adam claims that Bob attacked him first and he only attacked Bob out of self-defense. The Italian Penal Code provides two main articles that are relevant for this situation: art. 52 which states that a person is not liable if they commit an attack to defend themselves from an imminent threat, if the defense is proportionate to the threat; and art. 55 which explains that self-defense is not justified if the defense exceeds the attack. 

The arguments could be mapped as:
\begin{itemize}
    \item $B1$: Adam is liable for assaulting Bob.
    \item $A1$: Adam’s attack is justified by self-defense.
    \item $A2$: Bob threatened Adam with a weapon.
    \item $A3$: Adam was in immediate danger of aggression.
    \item $B2$: Adam used excessive force.
    \item $B3$: Bob tried to walk away but Adam continued to attack him.
    \item $B4$: Bob’s injuries are very serious while Adam has minor scratches.
\end{itemize}

The acceptance conditions for each argument are the following:
\begin{itemize}
    \item $AC(B1) = (B2 \wedge \lnot A3) \wedge \lnot A1$
    \item $AC(A1) = (A3 \wedge \lnot B2) \vee (A3 \wedge \lnot B1)$
    \item $AC(A2)$ is externally defined
    \item $AC(A3) = A2 \wedge \lnot B3$
    \item $AC(B2) = B3 \vee B4$
    \item $AC(B3)$ is externally defined
    \item $AC(B4)$ is externally defined
\end{itemize}

The acceptance or rejection of arguments $A2$, $B3$ and $B4$ cannot be easily inferred from the other arguments, so we assume they are verified by external proof like testimonies, surveillance videos or physical evidence of Bob’s and Adam’s wounds.

\begin{figure}[!t]
    \centering
    \includegraphics[width=0.7\linewidth]{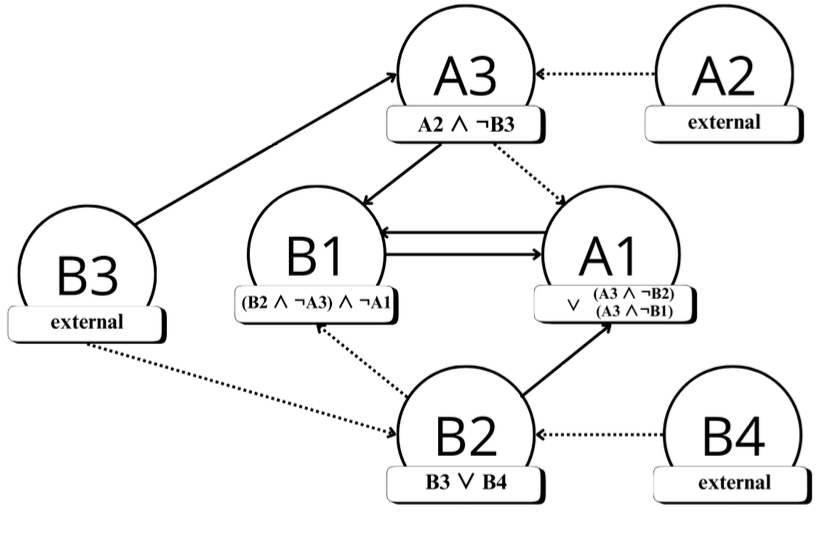}
    \caption{Graph of the Abstract Dialectical Framework used in the fourth
example.}
    \label{fig:adf}
\end{figure}

Fig.~\ref{fig:adf} shows the graph of the ADF for this example, where every argument is associated with its acceptance conditions. The attack relations are identified by solid line arcs and they represent the negation of an argument in the acceptance condition, for example the attack $(B2,A1)$ symbolizes $\lnot B2$ that is part of the acceptance condition of $A1$. There are also support relations in the graph, which
are identified by dotted line arcs and symbolize the acceptance of an argument in the acceptance condition, for example the support $(A3,A1)$ represents $A3$ that is part of the acceptance condition of $A1$.

For applying ADF semantics we start with the externally defined arguments
$A2$, $B3$ and $B4$, which are assumed accepted. Since $B3$ and $B4$ (Bob tried to walk away and Bob’s injuries are much worse than Adam’s) are both accepted, $B2$ (Adam used excessive force) is also accepted. In order for $B1$ (Adam is liable for assault) to be accepted $B2$ must be accepted (which it is), $A3$ must be rejected and $A1$ must be rejected. $A3$ is indeed rejected because even if it is supported by $A2$ (which is accepted), $B3$ is not rejected: so although Bob threatened Adam, Adam was not in immediate danger since Bob tried to walk away. The rejection of $A3$ also causes the rejection of A1, since both its acceptance conditions needed $A3$ to be accepted. So ultimately, $B1$ is justified since its acceptance conditions are met and Adam is liable for assault because even though Bob threatened him first, he went too far and kept attacking him after the threat was over, exceeding the limits imposed for self-defense. 

\begin{landscape}
\begin{table}[htbp]
\centering
\caption{Comparison of explanation approaches in AI and Law. The table highlights the distinctive contribution of computational argumentation frameworks to legal explainability.}
\label{tab:comparison_explanation_methods}
\begin{tabularx}{\linewidth}{
    >{\raggedright\arraybackslash}p{2.5cm}
    >{\raggedright\arraybackslash}p{3.2cm}
    >{\raggedright\arraybackslash}p{3cm}
    >{\raggedright\arraybackslash}p{3.2cm}
    >{\raggedright\arraybackslash}p{3.2cm}
    >{\raggedright\arraybackslash}p{3.2cm}
}
\hline
\textbf{Framework} & \textbf{Core Structure} & \textbf{Explainability Dimension} & \textbf{Strengths} & \textbf{Limitations} & \textbf{Legal Applicability} \\
\hline
\textbf{Example-based} & Instance-to-instance comparison & Intuitive, contrastive & Human-friendly and easy to grasp & Limited generalization capacity & Useful for case-based legal reasoning \\
\textbf{Rule-based} & Logical or symbolic inference chains & Transparency, traceability & Deterministic and logically interpretable & Rigid and domain-dependent & Well aligned with statutory or codified reasoning \\
\textbf{Hybrid} & Combination of symbolic and sub-symbolic models & Multi-level explanation & Balances accuracy and interpretability & Complex to integrate and validate & Suitable for AI-assisted legal prediction tasks \\
\textbf{Argumentation-based} & Argument–attack/defence relations & Contestability and justification & Models reasoning dynamics; supports interactive and contestable explanations & Requires formal modelling and computational resources & Ideal for GDPR/AIA-compliant explainable and contestable AI systems \\
\hline
\end{tabularx}
\end{table}
\end{landscape}

\section{Discussion}\label{sec:discussion}

Each Argumentation Framework reviewed offered unique insights into how legal arguments interact and influence one another, also demonstrating that both simpler structures like AAFs and more complex ones such as ADFs have high expressive capacity and potential to model legal problems, predict their outcome and justify it through contestable explanations.

To better understand the characteristics of the different types of explanation methods presented in the previous section, it is useful to compare their strengths and weaknesses in the legal system. Table~\ref{tab:comparison_explanation_methods} summarizes the main discussion points that will be developed in the following paragraphs.

Example-based methods are very intuitive, as they present concrete examples
that are either similar or contrastive to the case that is being analyzed, and flexible, as they can generalize past cases and be able to formulate a decision even for uncertain or incomplete situations. Since these explanations rely on precedents and analogies, they strongly resemble legal reasoning in common law systems, so they could cause confusion if applied to civil law tradition, also because the underlying rules behind a decision are not explicitly stated.

In contrast, rule-based explanations are more transparent because they
explicitly state the principle that caused a certain outcome, but they are less flexible than example-based methods and fail to produce a decision for new cases that do not fall under any existing rule. Moreover, a lot of effort is required to formalize rules, both from statue law as it often uses generic words and sometimes there are even contradictions, and precedent cases, as capturing the legal reasoning process is not easy.

Hybrid explanations should ideally merge the best features of the previous two methods and fill in their shortcomings by providing more balanced explanations with both concrete examples and explicit reasoning, but they are more difficult to implement as combining two distinct methodologies is challenging. 

Argument-based methods align with legal reasoning, as both use elements like claims, warrants and rebuttals for explaining the reasoning process that led to a certain conclusion. Arguments even capture the defeasible nature of legal claims, as they can be challenged and revised, allowing multiple perspectives and not believing in an absolute truth, but in an appropriate decision for a certain case instead. However, it is important to consider that structing an argument and evaluating its interactions with other arguments is not straightforward and requires an attentive implementation of the model.  

In order to successfully explain decisions or actions of an AI system to a human, it is necessary to look at how humans explain to each other. Generating more human-like explanations is the key to achieve understandable and satisfactory XAI  models. However, it is also essential to consider the context in which these explanations are given: for
example, legal decisions do not simply require a justification but a more complex answer that allows persuasion, refutation and deliberation. In most court procedures, judges do not just analyze the facts of the case and pronounce a sentence, but they also listen to and consider the arguments presented by the opposing parties, who intend to persuade the judge that their side is right. The judge then decides which party to favor and provides a justification for the sentence, but that wouldn’t have been enough if the argumentation part were to be skipped. The main point is that argumentation is an interactive process and not just a static explanation: if people were given an inflexible solution---even if thoroughly explained---without having the possibility to defend themselves by presenting counterarguments or exceptions, the process would not only feel unsatisfactory but also unfair. 

Moreover, the defeasible nature of argumentation is also ideal to highlight
other important parts of legal decision-making: claims can be challenged,
exceptions can be pointed out, and new information can lead to revised
conclusions. It is necessary to provide both descriptive and persuasive
explanations for XAI systems to align with the legal decision-making process: descriptive explanations clarify the functioning of the AI model and its reasoning in reaching the outcome, which could include applying rules or citing precedents, while persuasive explanations construct arguments and evaluate their relationships, building a structured debate. 

Argumentation Frameworks (AFs), which can be used to structure explanations around components like warrants and rebuttals, can contribute to replicate legal reasoning and make argumentation stand out among the different explanation techniques reviewed in the previous sections. Additionally, AFs can also effectively represent the burdens of proof and their dynamic nature, which are the core of legal disputes as the debate is often about whether a claim is sufficiently supported by evidence. The degree of justification needed for proving a certain claim in a specific context can also be determined by the AI system while arguing.

A major concern with argumentation-based explanations is their complexity: as opposed to simple example-based or rule-based methods, AI systems are required to construct arguments and analyze conflicting data. However, this complexity is also a reflection of legal reasoning: it is only natural for a complex decision-making process to be explained by a complex system. Legal decisions are also rarely straightforward, and explanations should avoid oversimplifying them, but with XAI there is a possibility to provide explanations with different levels of complexity. This is a big advantage as they can adapt to the knowledge of the user: for example, a judge utilizing an AI system as a decision-support tool might require a detailed explanation, whereas a layperson could benefit more from a simplified version.

Another challenge is ensuring that AI-generated arguments remain aligned
with legal standards and do not contain biases or errors in their reasoning, but this is a challenge for all explainability methods. Argumentation-based approaches actually have an advantage as it is easier to identify reasoning errors thanks to their transparent structure, and they also have the potential of making significant improvements in the reasoning capabilities of AI models by forcing the model to reason in a structured manner, since it has to provide structured explanations.

Additionally, the adoption of argumentation-based explanations for AI models could improve trust and acceptance of AI in the legal field. Trust is a fundamental issue in this context because automated decision-making has great power as it directly impacts individuals’ rights and freedom. This authority that AI would have is also recognized by the new Artificial Intelligence Act (AIA)~\cite{eu2024aiact}, which classifies AI systems used in the legal field as ``high-risk''. Explanations consisting in argumentation align with many of AIA’s requirements for high risk
systems, like the interactive reasoning that gives to the user the possibility of challenging the decision, as only providing a non-questionable justification would be insufficient to meet these regulatory requirements. Moreover, the AIA supports transparency by design, highlighting the importance of AI systems generating explanations as part of their decision-making process and, as we said before, argumentation forces the AI model to reason in a structured manner in order to
provide the structured arguments that are part of the explanation.
Argumentation-based explanations align well also with the other existing legal frameworks, such as the GDPR~\cite{eu2016gdpr} and the Recommendation on the Ethics of Artificial Intelligence~\cite{unesco2022ethics}, as they all stress the need for AI answers to be understandable and contestable, which are attributes that are naturally present in argumentation. Meeting AIA’s requirements for explainability is a particularly good sign because it means argumentation is consistent with the direction that AI regulations are moving towards. 

While it remains uncertain whether AI could ever be able to replace the jury, there is significant potential for AI to assist in the legal domain: with the right approach to explainability, AI can help speeding up the
judicial proceedings, improving accuracy and consistency by minimizing human error, reducing costs and lowering administrative burdens.

\section{Other Relevant Works}\label{sec:related}


The concept of XAI in the legal context is not new but the interest in this topic spread widely in the last decade, with the introduction of the GDPR~\cite{eu2016gdpr} in 2016. It will only continue to increase as the AI Act~\cite{eu2024aiact} will gradually come into force over the next years.

An overview of the most innovative works in this field is needed to better
understand the XAI framework, starting with a notable example-based model, the one proposed by Artelt et al.~\cite{artelt2023idk}, which is based on the idea that, in case of ambiguous situations, the system should abstain itself from giving a low-confidence answer and leave the decision to a human expert instead. It is a valuable approach that can help overcome the shortcomings of current explanation methods, but the main goal is to improve explanations so that the intervention of a human expert is not needed anymore.

An interesting research for argumentation-based explanations is that of Flouris et al.~\cite{flouris2023theoretical} which introduced Abstract Argumentation Frameworks with Domain Assignments (AAFDs) to overcome the limitations of argumentation frameworks in handling sets of entities and their properties. AAFDs assign a domain of application to each argument, allowing for a more refined evaluation of arguments by defining a set of entities for which the argument is applicable, rather than treating arguments as universally valid or invalid. The possibility of having a partial acceptance of an argument transforms argumentations with classical AFs into more flexible and precise debates, which resemble real-life discussions.

Fazzinga et al.~(2023)~\cite{fazzinga2023who} also contributed to improving argumentation-based explanations by introducing a new ``who said what'' relation, which is a function that maps each agent to a set of arguments that they claimed. This function is useful when agents are assigned a degree which represents their trustworthiness, so the strength of the claim depends by which agent proposed it.

Some of the most important AI and law projects are the HYPO project of
Ashley~\cite{ashley1990modeling}, which is a case-based reasoning system that modelled adversarial arguments in three separate steps: first the case is cited, then it is contested by the other party, and lastly the original party is has a chance to rebut.

In the subsequent years, the HYPO project split into CABERET~\cite{skalak1992arguments}, which aimed to extend case-based reasoning to formulate legal decisions, and CATO~\cite{aleven1997teaching}, which focused on using case-based reasoning for assisting law students in recognizing the distinguishing factors between precedents. These two separate paths represent an ongoing debate about what should AI pursue in the legal domain, either substituting judges or being restricted to a supporting tool. The first
hypothesis is the most controversial, but at the same time more appealing for many reasons, one being the judicial time would decrease if the decisions were taken by an automated system.

\section{Conclusions and Future Works}\label{conclusion}

In this paper we addressed the critical challenge of ensuring transparency of AI systems used in legal decision-making, which are often called black boxes because of their opacity, by providing explanations about the reasoning behind the outcome they predicted.

The legal framework---mainly consisting of the
General Data Protection Regulation (GDPR) and more recently the Artificial
Intelligence Act (AIA)---is clear about the right to explanation, especially in critical situations like the legal environment where a decision can have a big impact on people’s lives, like limiting their freedom. According to these regulations, the provided explanations must be comprehensible and contestable, thus we analyzed the most known explainability techniques that were studied and employed for AI models to find the most suitable one for the legal context.

Example-based explanations are based on analogies, hence they are very
intuitive and align well with the Common Law tradition, but they are not clear about the reasons that the system used to reach a decision. Rule-based explanations, on the other hand, explicitly state the legal principle that caused a certain outcome, but they are not flexible enough as they struggle to handle cases that do not fit into predefined rules. Additionally, both of these explanation methods cannot be challenged, so they do not abide by the regulations which demand contestable explanations. 

Argumentation-based explanations, in contrast, are both understandable and contestable, and because of that they emerge as the most robust explainability technique, especially for AI systems used in judicial decision-making as they inherently mirror legal disputes by structuring reasoning about claims, warrants and exceptions which, if appropriate, can modify the previous decision. However, there are still many challenges that should be addressed in future works, like ensuring that AI-generated arguments are free of bias and aligned with existing laws, but there is hope that with the right approach to explainability, AI systems will be able to assist in speeding up judicial proceedings, improving accuracy and reducing costs while maintaining the fairness and legitimacy of legal decisions.

This paper presents a first analysis of XAI in the legal field so further
exploration remains open, like examining the presence of biases in training data and possible ways to remove or mitigate them, so that argumentation-based models more would be more precise and applicable to judicial decision-making.

Future works could also test whether AI systems can just be used as supportive tools for judges and other legal professionals, or if they could eventually fully replace human judgement. The most feasible option is the first one, but even if AI only had an assistive role it would still revolutionize the legal system, so there should be research and empirical studies to try to understand how the dynamics in the judicial processes would change if AI systems were to be adopted on a large scale, for example how the opinion of people would change and if they would still
trust the legal system or feel unsafe. This type of research could also provide practical insights on how to improve AI’s usability and effectiveness based on how legal professionals and laypeople interact with it, and possibly in preparing for the transition to employing AI in this context while maintaining the authority and fairness of law. In the future, researchers should also ensure that AI systems comply with evolving ethical and regulatory standards, especially in relation to the Artificial Intelligence Act, which will introduce the CE marking for AI models in the European Union. A model which does not meet the standards and does not have the CE marking will not be allowed to be used in any EU member state, so all AI developers should pay attention to these standards to prevent their models from being limited. 

By examining various XAI techniques and assessing their suitability in the
legal environment, this paper also contributes to the field of Explainable AI with the hypothesis that, among the multiple explainability methods that have been developed, argumentation-based approaches are the most robust to ensure the transparency and trustworthiness of AI models employed in the legal system. In fact, as a result of their structured and logical approach to justifying decisions, arguments are best aligned with legal reasoning and with the evolving legal standards, like the Artificial Intelligence Act, also thanks to their interactive framework that allows for the possibility to challenge the decisions. Despite this, the research also identifies ongoing challenges, like ensuring that the arguments generated by AI models are free of biases and based on existing laws, so further research and testing are needed in order to be able to implement XAI models in judicial processes in the future.




\end{document}